\documentclass[conference]{IEEEtran}
\IEEEoverridecommandlockouts
\usepackage{cite}
\usepackage{amsmath,amssymb,amsfonts}
\usepackage{graphicx}
\usepackage{textcomp}
\usepackage{xcolor}
\usepackage{xspace}
\usepackage{xurl}
\def\BibTeX{{\rm B\kern-.05em{\sc i\kern-.025em b}\kern-.08em
    T\kern-.1667em\lower.7ex\hbox{E}\kern-.125emX}}
\usepackage{mathtools}
 
\usepackage{multirow}
\usepackage{adjustbox}
\usepackage{tikz}
\usetikzlibrary{positioning}
\usepackage{hyperref}
\usepackage{caption}
\usepackage[french,boxed,vlined,linesnumbered,inoutnumbered,rightnl,algo2e]{algorithm2e}
\usepackage{algorithm}
\usepackage{mwe}
\usepackage{stfloats}
\usepackage{pifont}

\IEEEoverridecommandlockouts
\IEEEpubid{\makebox[\columnwidth]{979-8-3315-3477-6/25/\$31.00~\copyright2025 IEEE\hfill} \hspace{\columnsep}\makebox[\columnwidth]{ }}

\begin{document}
\title{Automating Versatile Time-Series Analysis with Tiny Transformers on Embedded FPGAs\thanks{The authors gratefully acknowledge the financial support provided by the Federal Ministry for Economic Affairs and Climate Action of Germany for the RIWWER project (01MD22007C).}}
\author{
    \IEEEauthorblockN{
    Tianheng Ling, 
    Chao Qian,  
    Lukas Johannes Haßler, and
    Gregor Schiele
    }
    \IEEEauthorblockA{
    Department of Intelligent Embedded Systems, University of Duisburg-Essen 
    }
}
\maketitle
\begin{abstract}
Transformer-based models have shown strong performance across diverse time-series tasks, but their deployment on resource-constrained devices remains challenging due to high memory and computational demand. While prior work targeting Microcontroller Units (MCUs) has explored hardware-specific optimizations, such approaches are often task-specific and limited to 8-bit fixed-point precision. Field-Programmable Gate Arrays (FPGAs) offer greater flexibility, enabling fine-grained control over data precision and architecture. However, existing FPGA-based deployments of Transformers for time-series analysis typically focus on high-density platforms with manual configuration.
This paper presents a unified and fully automated deployment framework for Tiny Transformers on embedded FPGAs. Our framework supports a compact encoder-only Transformer architecture across three representative time-series tasks (forecasting, classification, and anomaly detection). It combines quantization-aware training (down to 4 bits), hardware-aware hyperparameter search using Optuna, and automatic VHDL generation for seamless deployment.
We evaluate our framework on six public datasets across two embedded FPGA platforms. Results show that our framework produces integer-only, task-specific Transformer accelerators achieving as low as 0.033 mJ per inference with millisecond latency on AMD Spartan-7, while also providing insights into deployment feasibility on Lattice iCE40. All source code will be released at: \url{https://github.com/tianheng-ling/TinyTransformer4TS}.

\end{abstract}
\begin{IEEEkeywords}
Time-Series Analysis, Embedded AI, Tiny Transformers, Model Quantization, FPGAs Acceleration
\end{IEEEkeywords}


\section{Introduction}
\label{sec:introduction} 

The rapid growth of the Internet of Things (IoT), embedded sensing, and edge intelligence has created a strong demand for deploying Deep Learning (DL) models directly on resource-constrained hardware~\cite{wang2025empowering}. On-device inference enables fast responses, reduces dependence on Cloud connectivity, and protects privacy by processing data near the source~\cite{simuni2024edge}. These properties are essential for applications such as anomaly detection in smart infrastructure, activity recognition in wearables, and environmental forecasting with smart sensors.

In recent years, Transformer-based architectures have gained widespread popularity in natural language processing and computer vision, and more recently in time-series analysis~\cite{wen2023transformer}. Their ability to capture long-range dependencies makes them particularly powerful in time-series forecasting, classification, and anomaly detection tasks. However, their high memory footprint and the complexity of self-attention severely limit their deployability on constrained edge hardware~\cite{tabani2021improving}.

Early efforts to bring time-series Transformers to the extreme edge have focused on Microcontroller Units (MCUs), leveraging such as fused kernels and handcrafted scheduling~\cite{jung2024optimizing}. While effective, these approaches are often platform-specific and limited to 8-bit fixed-point arithmetic. In contrast, Field-Programmable Gate Arrays (FPGAs) offer a compelling alternative~\cite{yan2024survey}, enabling fine-grained control over arithmetic precision and hardware architecture. However, existing FPGA-based Transformer implementations often rely on manual design, target mid- to high-density platforms, and lack generalizability across tasks and datasets.

To address these gaps, we propose a unified and fully automated deployment framework for Tiny Transformers targeting embedded FPGAs. Our approach integrates model quantization, hardware-aware optimization, and deployable RTL code generation to support real-world time-series analysis at the edge. The contributions of this work are:
\begin{itemize}
    \item We develop an encoder-only Transformer architecture across three representative time-series tasks (single-step forecasting, classification, and threshold-based anomaly detection) without task-specific modifications.

    \item We integrate hardware-aware optimization through quantization-aware training (down to 4 bits), Optuna-based hyperparameter search, and deployability filtering tailored to embedded FPGA resource constraints.

    \item We automate VHDL code generation and synthesis via modular templates, enabling low-power deployment on AMD Spartan-7 FPGA and analyzing feasibility on Lattice iCE40 FPGA.

    \item We evaluate our framework across six public time-series datasets, achieving millisecond latency and energy consumption as low as 0.033 mJ per inference.

\end{itemize}

The remainder of this paper is organized as follows: 
Section~\ref{sec:related_work} reviews related work.
Section~\ref{sec:framework} details the proposed deployment framework.
Section~\ref{sec:results_eval} presents experimental results across three time-series tasks.
Section~\ref{sec:conclusion_future_work} concludes the paper and reflects on limitations that motivate future research.

\section{Related Work}
\label{sec:related_work}

This section reviews efforts to deploy Transformer models for time-series analysis on resource-constrained platforms, focusing on MCUs and FPGAs.

\subsection{Deploying Transformers on MCUs}

Transformer-based models have increasingly been adapted to MCU platforms for time-series tasks, leveraging the affordability and mature deployment ecosystem of MCUs.
Becnel et al.~\cite{becnel2022tiny} introduced T$^3$, a compact Encoder-only Transformer for multivariate forecasting, achieving real-time inference on an ESP32 MCU using full 8-bit quantization via TensorFlow Lite. However, Transformer-based anomaly detection on MCUs remains limited, with many studies still favoring simpler models such as Multilayer Perceptrons (MLPs), Convolutional Neural Networks (CNNs), or Long Short-Term Memory (LSTMs), as claimed in a systematic study~\cite{trilles2024anomaly}. 

Time-series classification has attracted significant attention. Early works such as Burrello et al.~\cite{burrello2021microcontroller,burrello2022bioformers} proposed lightweight attention blocks for sEMG-based gesture recognition. Subsequent studies by Busia et al.~\cite{busia2024reducing,busia2024tiny} developed task-specific Transformers for EEG and ECG classification on GAP9-class MCUs. Jung et al.~\cite{jung2024optimizing} unified these developments, introducing a deployment framework with fused-weight attention, optimized scheduling, and cross-platform kernel libraries for MCUs based on RISC-V and ARM architectures. While these studies demonstrate impressive efficiency, they are often tightly coupled to specific instruction sets, rely on fixed operator scheduling, and typically operate under 8-bit fixed-point precision. 


\subsection{Deploying Transformers on FPGAs}

Compared to MCUs, FPGAs offer fine-grained control over both architecture and precision, making them well-suited for ultra-efficient deployment. However, most FPGA-based Transformer deployments to date focus on mid- or high-density platforms.
For instance, Yu et al.~\cite{yu2024dual} proposed a CNN-Transformer hybrid architecture for multivariate forecasting and classification, targeting Xilinx Ultra96V2 FPGA.
Sobakinskikh et al.~\cite{sobakinskikh2024optimizing} optimized Transformer inference for financial outlier detection on PYNQ-Z2 board featuring the ZYNQ XC7Z020-1CLG400C SoC. In addition, Wang et al.~\cite{wang2024mr} proposed MR-Transformer for modulation recognition in communication systems, using matrix tiling and reuse strategies on Xilinx Zynq UltraScale+ MPSoCs development platform with XCZU3EG FPGA. 

Efforts targeting truly resource-constrained FPGAs have only recently emerged. Ling et al.~\cite{ling2024integer} demonstrated the first fully integer-only Transformer on AMD Spartan-7 for time-series forecasting, employing 4-bit quantization via Quantization-aware Training (QAT) for significant energy savings. A follow-up study~\cite{ling2024resource} introduced mixed-precision quantization with resource estimation to improve deployability.

Nonetheless, these works remain limited in scope. They typically address only a specific task, require manual model configuration, and lack hardware-aware automation.
To the best of our knowledge, our work is the first to enable multi-task, quantized Transformer deployment with full automation, including training, hardware-aware search, and VHDL synthesis, targeting multiple embedded FPGA platforms.

\section{Deployment Framework for Transformers}
\label{sec:framework}

This section provides an overview of our automated deployment framework and then details on each of its stages, from model design to hardware verification on embedded platforms.

\vspace{-5pt}
\begin{figure}[!htb]
    \centering
    \includegraphics[width=.85\columnwidth]{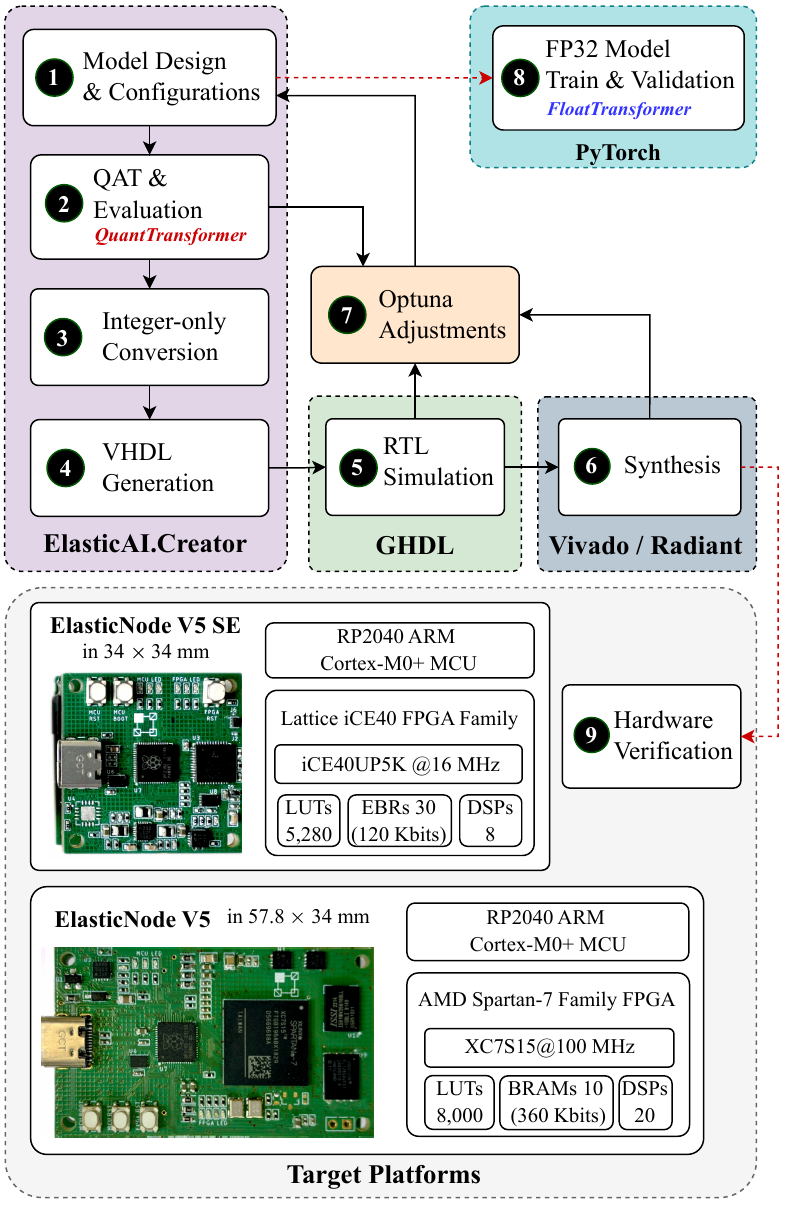}
    \caption{Visualization of deployment framework.}
    \label{fig:deployment_flow}
\vspace{-15pt}
\end{figure}


\subsection{Overview of Deployment Framework}

As shown in Figure~\ref{fig:deployment_flow}, our framework begins by designing and configuring a \emph{QuantTransformer} model (\ding{183}) using quantization-compatible layer or module from an open-source \emph{ElasticAI.Creator}\footnote{\url{https://github.com/es-ude/elastic-ai.creator/tree/add-linear-quantization}} library~\cite{qian2023elasticai}. QAT is applied to calculate quantization parameters for later enabling integer-only inference. Once training converges, the model is exported to an integer format (\ding{184}) suitable for hardware mapping.
In the VHDL generation phase (\ding{185}), each quantized layer is translated to a RTL module with parameterized bitwidths and tensor dimensions. 
The design is first functionally verified using GHDL RTL simulation (\ding{186}) and then synthesized with Vivado (for AMD Spartan-7) or Radiant (for Lattice iCE40) (\ding{187}), generating post-synthesis reports on logic utilization, timing, and estimated power consumption.
Inspired by hardware-aware optimization in LOTTA~\cite{kress2024lotta}, we integrate an Optuna-driven hyperparameter search loop (\ding{188}). Each candidate configuration undergoes QAT, simulation, synthesis, and deployability filtering. Only configurations that pass hardware constraints proceed to final validation.
To establish a performance baseline, each selected configuration is also evaluated using a PyTorch-based \emph{FloatTransformer} model (\ding{189}) trained and validated on the same dataset in full-precision (FP32). The final quantized models are then deployed on real hardware for end-to-end testing and performance validation (\ding{190}).
We target two embedded FPGA platforms: ElasticNode V5 (AMD Spartan-7 XC7S15 @100\,MHz, 8,000 LUTs, 20 DSPs, 10 BRAMs), and ElasticNode V5 SE (Lattice iCE40UP5K @16\,MHz, 5,280 LUTs, 8 DSPs, 30 ERBs). Both platforms integrate an RP2040 Cortex-M0+ MCU coordinating data acquisition and managing inference requests to the FPGA~\cite{ling2024configurable}.


\begin{figure}[!htb]
\vspace{-5pt}
    \centering
    \includegraphics[width=.82\columnwidth]{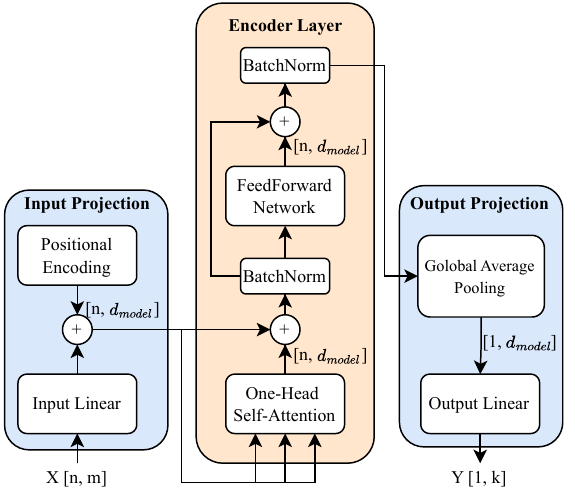}
    \caption{The architecture of the Transformer model.}
    \label{fig:transformer_model}
\vspace{-15pt}
\end{figure}

\subsection{Tiny Transformer Architecture \ding{182}\ding{189}}

We adopt a compact encoder-only Transformer as a unified backbone for three time-series tasks, which balances modeling capacity and deployment simplicity. The design builds on prior work~\cite{ling2024integer}, with minor adjustments to ensure task compatibility. As illustrated in Figure~\ref{fig:transformer_model}, the architecture consists of three main components: 
(1) Input Projection: The input sequence $X \in \mathbb{R}^{n \times m}$, with $n$ time steps and $m$ features, is projected into a $d_\text{model}$-dimensional latent space via a linear layer, followed by positional encoding.
(2) Encoder Block: A single-layer encoder includes one-head Self-Attention(OHSA) and a Feedforward Network (FFN), each with residual connections and Batch Normalization (BatchNorm). The feedforward hidden size is set to $4 \times d_\text{model}$.
(3) Output Projection: The encoder output is aggregated using Global Average Pooling and passed through a final linear layer to produce task-specific output $Y \in \mathbb{R}^{1 \times k}$, where $k$ depends on the task (e.g., number of classes or predicted variables).
The same model architecture is reused across all tasks without structural changes. Task-specific components such as loss functions or anomaly thresholds are handled externally.

\subsection{Quantization with ElasticAI.Creator \ding{183}\ding{184}}

To support inference on embedded FPGAs, we apply fully integer-only quantization via QAT. Following~\cite{ling2024integer}, all model components (including weights, activations, inputs, and outputs) are quantized during training, allowing deployment with integer-only arithmetic. Our implementation uses uniform asymmetric quantization for most tensors to enhance accuracy, while biases and BatchNorm statistics are quantized symmetrically to reduce computational complexity during inference. The quantization scale $S$ and zero-point $Z$ are computed from each tensor’s dynamic range $[\alpha, \beta]$, as shown in Equation \ref{eq_asymmetric_signed_zero_point}. The quantization process is defined as Equation \ref{eq:quantization_mapping}, while the corresponding dequantization is given by Equation \ref{eq:dequantization_mapping}. 

\begin{footnotesize}
\begin{align}
S &= \frac{\beta - \alpha}{2^b - 1} \label{eq_asymmetric_signed_scale} \\
Z &\approx \text{clamp}\left(\text{round}((2^{b-1} - 1) - \frac{\beta}{S}),\; -2^{b-1},\; 2^{b-1}-1\right)
\label{eq_asymmetric_signed_zero_point}
\end{align}
\begin{equation}
X \mapsto X_q \approx \text{clamp}\left(\text{round}\left(\frac{X}{S}\right) + Z,\; -2^{b-1},\; 2^{b-1}-1\right)
\label{eq:quantization_mapping}
\end{equation}
\begin{equation}
X_q \mapsto X' = S \cdot (X_q - Z)
\label{eq:dequantization_mapping}
\end{equation}
\end{footnotesize}

Unlike MCU-focused deployment that applies  8-bit fixed-point quantization~\cite{jung2024optimizing}, our framework supports configurable bitwidths $b$ (as low as 4 bits), enabling finer trade-offs between accuracy and hardware efficiency. Each module in the resulting \emph{QuantTransformer} supports a dual-mode interface. During QAT, the \texttt{forward()} function simulates quantized behavior with gradient support. For deployment, the \texttt{int\_forward()} function executes deterministic integer-only arithmetic aligned with the generated VHDL implementation.

\subsection{VHDL Generation via Modular Templates \ding{185}}

Each quantized module in the \emph{QuantTransformer} is mapped to a reusable, parameterized RTL block implemented in VHDL. These hardware templates support configurable bitwidths, quantization parameters and tensor shapes, and are accompanied by corresponding testbenches for functional simulation. ElasticAI.Creator exposes a unified \texttt{design()} interface that injects specific parameters into the corresponding VHDL code, enabling automatic generation of synthesizable hardware components. For example, the quantized \texttt{Linear} layer is compiled into a integer-only matrix multiplier with shift-based scaling.
Moreover, platform-specific wrappers and constraints are inserted during the VHDL generation. Vivado-compatible timing and placement directives are used for AMD Spartan-7, while Radiant-specific constraint files are generated for Lattice iCE40. Top-level integration logic and firmware stubs are also auto-generated to produce synthesizable hardware designs ready for deployment.

\subsection{RTL Simulation and FPGA Synthesis \ding{186}\ding{187}}

Following VHDL generation, each design is first validated through RTL simulation using GHDL~(\ding{187}), ensuring cycle-accurate correctness and consistent integer-only behavior. Upon passing simulation, the design is synthesized using vendor-specific toolchains, Vivado 2019.2 for AMD Spartan-7 and Radiant 2023.2 for Lattice iCE40~(\ding{188}).
The synthesis process yields detailed reports on resource utilization (LUTs, BRAMs/EBRs, DSPs) and estimated power consumption. These metrics are automatically parsed and logged for downstream use in hardware-aware optimization. 

\vspace{-5pt}
\subsection{Hardware-Aware Optimization \ding{188}}

To efficiently guide the selection of model configurations that balance accuracy and deployment efficiency, we adopt an Optuna-based hyperparameter optimization strategy that integrates hardware feedback into each search trial. This addresses two limitations in our prior work~\cite{ling2024configurable}: (1) reliance on exhaustive grid search, which is computationally inefficient, and (2) the need for post hoc filtering of infeasible models. By embedding synthesis and simulation feedback directly into the optimization loop, the search process can jointly explore both algorithmic performance and hardware feasibility. 
Specially, at each trial $t$, Optuna samples a candidate configuration $\theta_t$ from the predefined search space. The corresponding \emph{QuantTransformer} is trained using QAT, and its validation loss $\texttt{Loss}_t$ is recorded. The trained model is then converted to integer-only format and translated into synthesizable VHDL. RTL simulation using GHDL verifies correctness and measures cycle-accurate latency $T_t$, while synthesis (Vivado or Radiant) reports resource usage and estimated power $P_t$. Trials that exceed the FPGA’s resource budget are discarded early. For deployable trials, we compute energy per inference as $E_t \!=\! P_t \times T_t$. Each trial returns a tuple $(\texttt{Loss}_t, \texttt{HW}_t)$, where $\texttt{HW}_t$ encapsulates hardware-level metrics (e.g., energy, latency), enabling multi-objective optimization driven by both algorithmic performance and deployment efficiency.

\section{Use Cases and Evaluations}
\label{sec:results_eval}

In this section, we evaluate our deployment framework across three time-series tasks, spanning six public datasets. All experiments target two embedded FPGA platforms.

\subsection{Experimental Setup}

Each experiment consists of 100 Optuna trials with the \texttt{NSGAIISampler} sampler, jointly minimizing validation loss and energy per inference. Only configurations that meet all FPGA resource constraints are retained for Pareto analysis. All models are trained with the following search space:
\begin{itemize}
    \item Quantization bitwidth: $b \in \{4, 6, 8\}$ 
    \item Batch size: $bs \in \{16, 32, \dots, 256\}$
    \item Learning rate: $lr \in [10^{-5}, 10^{-2}]$ (log-uniform)
    \item Embedding dimension: $d_\text{model} \in \{8, 16, \dots, 64\}$
\end{itemize}

Each trial runs for up to 100 epochs with early stopping (patience \!=\! 10). All experiments are conducted in Python 3.11 on an NVIDIA RTX 2080 SUPER GPU (CUDA 11.0). The Adam optimizer is used with default settings. Notably, we fix the operating frequency to 100\,MHz for AMD XC7S15 FPGA and 16\,MHz for Lattice iCE405UPK FPGA across all experiments for fair comparison.


\begin{table*}
\caption{Selected Transformer configurations for time-series tasks on AMD XC7S15 FPGA.}
\label{tab:optuna_best_trial}
\resizebox{\textwidth}{!}{
\centering
\begin{tabular}{|c|c|c|c|c|c|c|c|c|c|c|c|c|c|c|}
\hline

\multirow{2}{*}{Task} & \multirow{2}{*}{Datasets} & \multicolumn{5}{c|}{Configuration} & \multicolumn{2}{c|}{Accuracy Metrics$^\ddagger$} & \multirow{2}{*}{\begin{tabular}[c]{@{}c@{}}LUTs\\ (\%)\end{tabular}} & \multirow{2}{*}{\begin{tabular}[c]{@{}c@{}}BRAMs\\ (\%)\end{tabular}} & \multirow{2}{*}{\begin{tabular}[c]{@{}c@{}}DSPs\\ (\%)\end{tabular}} &
\multirow{2}{*}{\begin{tabular}[c]{@{}c@{}}Energy\\ (mJ)\end{tabular}} & \multirow{2}{*}{\begin{tabular}[c]{@{}c@{}}Power\\ (mW)$^*$\end{tabular}} & \multirow{2}{*}{\begin{tabular}[c]{@{}c@{}}Latency\\ (ms)$^{**}$\end{tabular}} \\ \cline{3-9}

& & b & bs & lr ($\times10^{-3})$  & $d_\text{model}$ & params$^\dagger$ & \multicolumn{1}{c|}{FP32} & INT & & & & &  & \\ \hline

\multirow{2}{*}{Forecasting} 
& PeMS & 6 & 208 & 6.357 & 16 & 3329 & 0.160 & 0.194 & 46.60 & 100 & 90 & 0.078 & 65.0 & 1.203 \\ \cline{2-15}
& AirU & 8 & 32 & 5.025 & 8 & 897 & 4.129 & 4.853 & 53.48 & 95 & 100 & 0.036
 & 64.0 & 0.570 \\ \hline

\multirow{2}{*}{Classification} 
& UCIHAR & 8 & 240 & 4.618 & 8 & 1006 & 0.858 & 0.824 & 53.64 & 100 & 90 & 0.067
 & 65.0 & 1.034 \\ \cline{2-15}
& WISDM & 6 & 48 & 1.257 & 40 & 20126 & 0.838 & 0.839 & 96.94 & 100 & 100 & 0.855 & 71.0 & 12.04 \\ \hline

\multirow{2}{*}{\begin{tabular}[c]{@{}c@{}}Anomaly\\ Detection\end{tabular}} 
& ALFA & 4 & 192 & 1.299 & 8 & 1106 & 0.923 & 0.889 & 35.55 & 100 & 65 & 0.033 & 62.0 & 0.527 \\ \cline{2-15}
& SKAB & 6 & 96 & 5.064 & 24 & 7465 & 0.766 & 0.765 & 58.95 & 100 & 95 & 0.154 & 68.0 & 2.261 \\ \hline

\multicolumn{15}{l}{$^\dagger$The number of model parameters.}\\
\multicolumn{15}{l}{$^\ddagger$Test RMSE for forecasting task, Test Accuracy for classification task, and Test F1 score for anomaly detection task.} \\
\multicolumn{15}{l}{$^*$ Power was estimated at a temperature of 28.0 $^{\circ}$C with a power deviation of 5.8\% compared to actual hardware measurements.} \\
\multicolumn{15}{l}{$^{**}$ The latency obtained from GHDL compared to actual hardware measurements varies by about 2\%.}
\end{tabular}
}
\vspace{-12pt}
\end{table*}


\subsection{Time-Series Forecasting}

We begin by evaluating our framework on two datasets using a 24-step sliding window for single-step ahead forecasting. 
PeMS\footnote{\url{https://doi.org/10.5281/zenodo.3939793}} is a univariate traffic flow dataset with 5-minute readings from over 11,000 sensors. Following~\cite{ling2024integer}, we extract data from sensor 4192.  
AirU\footnote{\url{https://dx.doi.org/10.21227/aeh2-a413}} is a multivariate air quality dataset containing 19,380 samples across seven features. Based on~\cite{becnel2022tiny}, we retain 15,258 valid samples and use ozone concentration as the prediction target, applying Min-Max normalization across all features.
During Optuna search, models are optimized using the Mean Squared Error (MSE) loss, and RMSE (calculated on denormalized targets and predictions) is used for final evaluation to ensure fair comparison with~\cite{ling2024integer}.

\begin{figure}[!htbp]
    \centering
    \includegraphics[width=0.49\columnwidth]{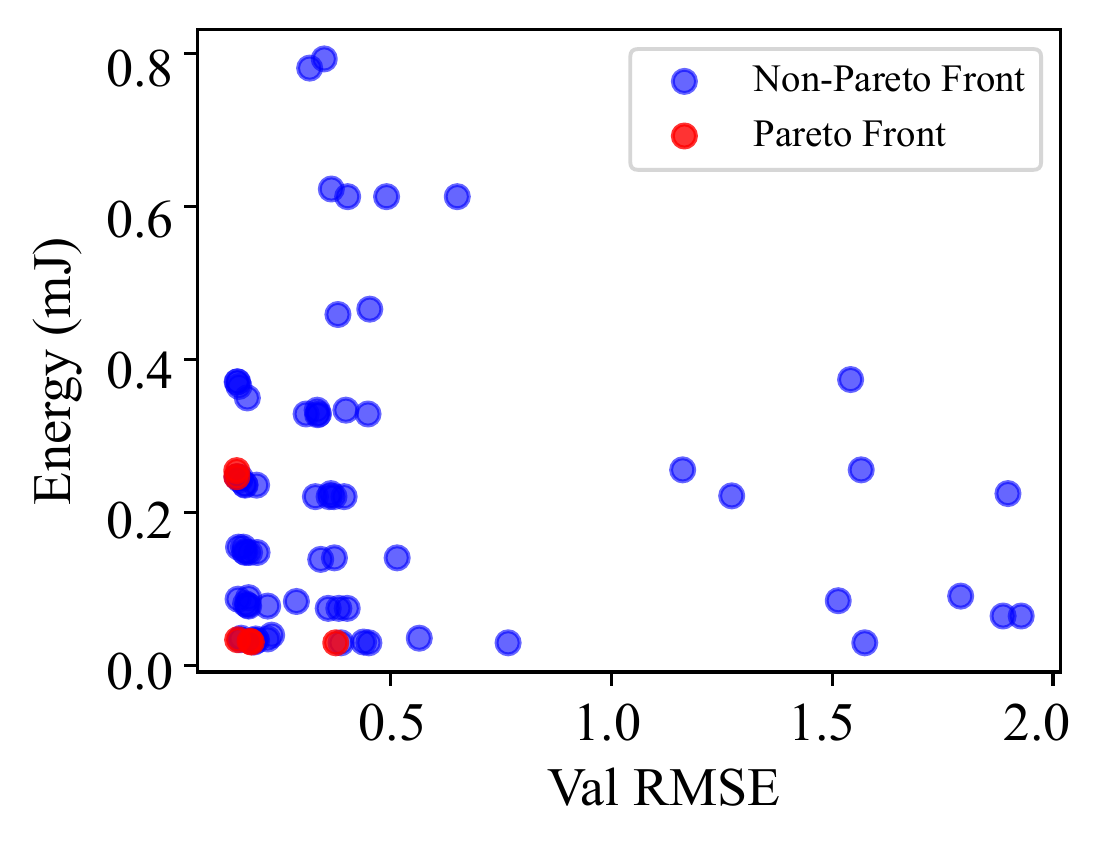}
    \includegraphics[width=0.49\columnwidth]{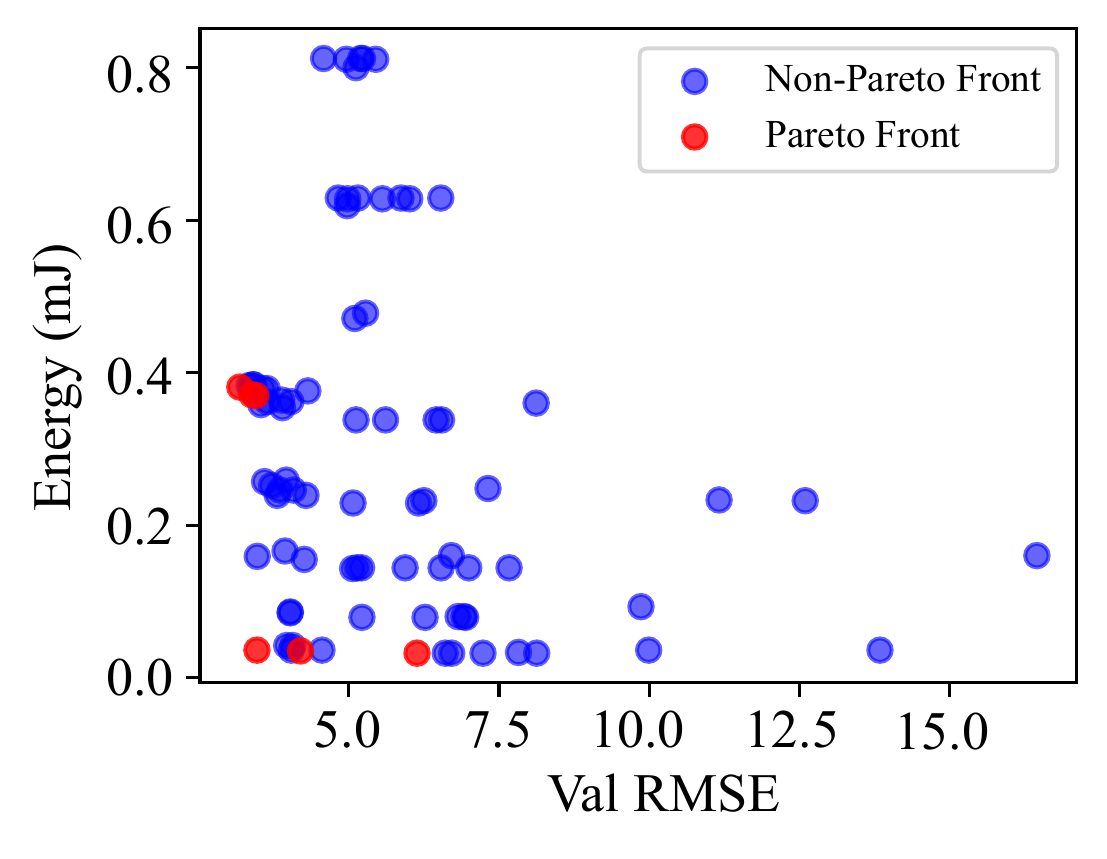}
    \caption{Pareto fronts on PeMS (left) and AirU (right).}
    \label{fig:forecasting_pareto_amd}
\end{figure}

Figure~\ref{fig:forecasting_pareto_amd} shows the Pareto fronts for PeMS and AirU datasets on XC7S15 FPGA, where each dot represents a valid configuration sampled via Optuna. Red markers denote Pareto-optimal trade-offs between denormalized RMSE and energy. 

\vspace{-8pt}
\vspace{-5pt}

Table~\ref{tab:optuna_best_trial} lists the Pareto-optimal configurations with the highest performance under integer-only quantization.
On the PeMS dataset, the selected model configuration ($d_\text{model}\!=\!16$, $bs\!=\!208$, $lr\!=\!6.357 \times 10^{-3}$) achieves a test RMSE of 0.194 under 6-bit quantization, representing a 21.25\% increase relative to the FP32 baseline (0.160), but a 2.41\% improvement over the same-sized 6-bit model reported in~\cite{ling2024integer}. This configuration demonstrates strong deployment efficiency, operating at 1.203 ms latency with only 0.078mJ energy per inference.
On the AirU dataset, the selected 8-bit configuration ($d_\text{model}\!=\!8$, $bs\!=\!32$) achieves an RMSE of 4.853, 17.53\% higher than the FP32 baseline (4.129). Compared to~\cite{ling2024integer}, where the same architecture yields an RMSE of 5.377, our model achieves a 9.75\% improvement. Furthermore, it surpasses their most energy-efficient variant (5.474 RMSE, 0.084 mJ, 1.33 ms) both in accuracy and efficiency, consuming only 0.036 mJ and 0.570 ms. When benchmarked against the MCU-based deployment in~\cite{becnel2022tiny} (176 ms latency, 4.048 mJ per inference, 5.44 RMSE), our FPGA-based implementation achieves 308$\times$ lower latency and 112$\times$ lower energy, while also improving accuracy.

For the Lattice iCE40UP5K platform, none of the 100 Optuna trials for either dataset yielded deployable designs. Even with the most compact configurations ($d_\text{model}\!=\!8$, 4-bit quantization), resource constraints were exceeded, where LUT usage surpassed the available budget by 16\%, all DSPs were exhausted, and EBRs utilization reached 97\%. This infeasibility was consistent across all tasks, so iCE40UP5K results are excluded from further evaluations. These findings highlight the challenges of deploying attention-based models on ultra-constrained FPGAs and motivate future work on architectural simplification or hybrid designs.

\subsection{Time-Series Classification}

\begin{figure}[ht]
\vspace{-6pt}
    \centering
    \includegraphics[width=0.49\columnwidth]{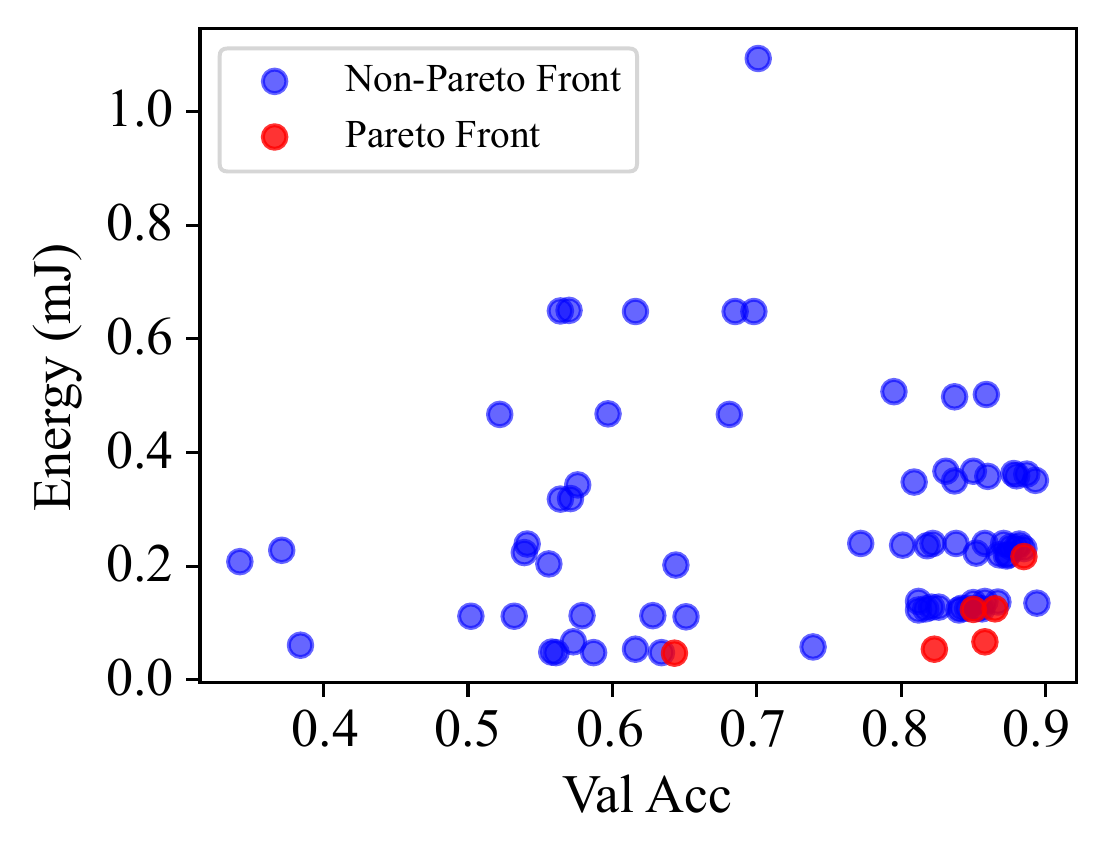}
    \includegraphics[width=0.49\columnwidth]{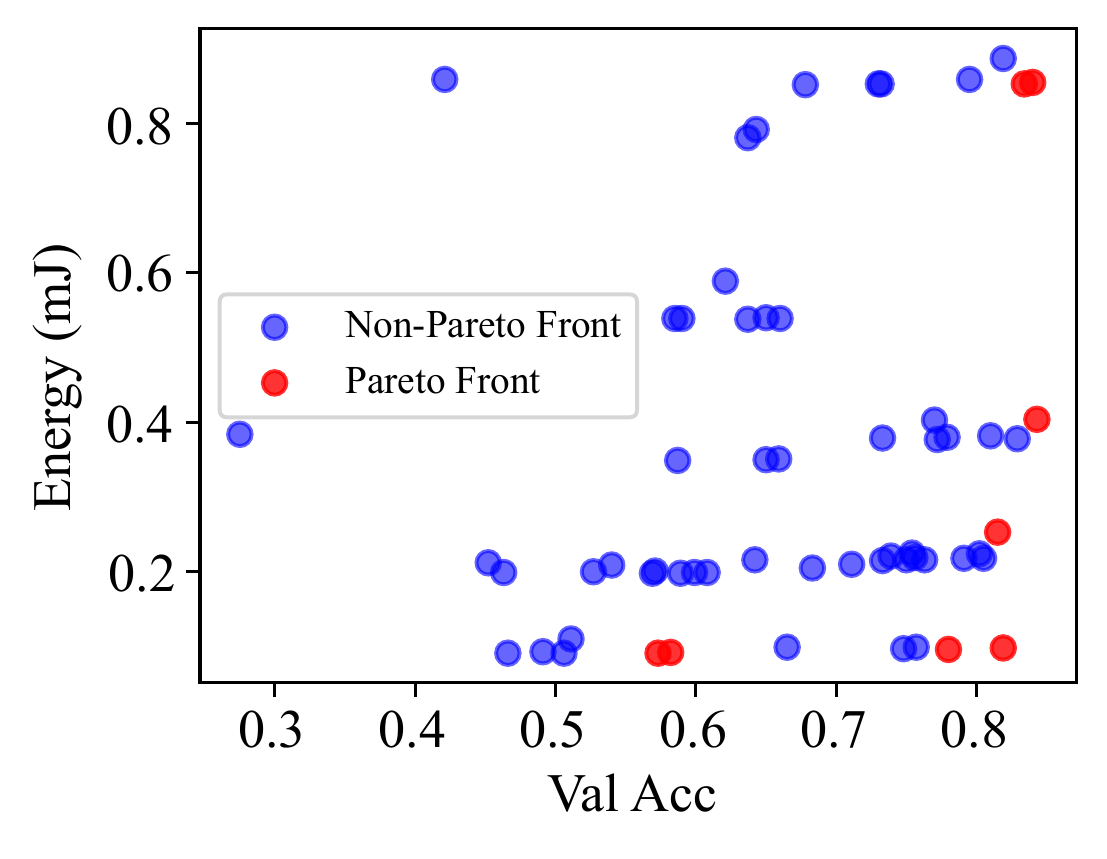}
    \caption{Pareto fronts on UCIHAR (left) and WISDM (right)}
    \label{fig:classification_pareto_amd}
\vspace{-6pt}
\end{figure}

We next evaluate our framework on two human activity recognition datasets to assess its suitability for classification tasks.
UCIHAR\footnote{\url{https://github.com/arijitiiest/UCI-Human-Activity-Recognition}} dataset comprises 9 sensor signals (accelerometer and gyroscope, 3-axis each) sampled at 50\,Hz. To reduce computational overhead, we apply a 4$\times$ downsampling, resulting in 32-step windows for 6-class classification. 
WISDM\footnote{\url{https://github.com/bartkowiaktomasz/har-wisdm-lstm-rnns}} dataset provides motion data with three channels. We downsample the input by 4$\times$ to obtain 50-step sequences, and limit experiments to 1600 samples per class due to constrained training resources.
During Optuna search, models are optimized using Cross-entropy loss and selected based on macro accuracy. Figure~\ref{fig:classification_pareto_amd} presents the Pareto fronts for both datasets, where red markers denote configurations that balance validation accuracy and energy on the XC7S15 FPGA.

As shown in Table~\ref{tab:optuna_best_trial}, on the UCIHAR dataset, the selected configuration ($d_\text{model}\!=\!8$, $bs\!=\!240$, $lr\!=\!4.618 \times 10^{-3}$) achieves a test accuracy of 0.824 under 8-bit quantization, representing only a 3.96\% degradation from the FP32 baseline (0.858). This modest drop suggests that classification tasks are inherently more resilient to quantization than forecasting. The model also demonstrates efficient deployment characteristics, operating at 1.034 ms latency, with an energy cost of 0.067 mJ per inference and an average power consumption of 65 mW.
In comparison, Samanta et al.~\cite{samanta2024optimizing} report a higher accuracy of 0.904 using a lightweight CNN with post-training quantization and 75\% input downsampling, targeting an MCU. While their model achieves an 8.85\% accuracy gain, the absence of details regarding quantization bitwidths and training protocols limits reproducibility. Despite the architectural simplicity of their model, our Transformer design achieves 7.44$\times$ lower latency (1.03 ms vs. 7.69 ms), highlighting its suitability for latency-sensitive applications. Although our energy consumption (0.067 mJ) is higher than theirs (0.041 mJ), this discrepancy could be attributed to the static power overhead of the XC7S15 FPGA (31 mW). These observations reinforce the importance of exploring deployment on ultra-low-power FPGAs such as the Lattice iCE40UP5K, where static power is negligible and further energy gains may be realized.

On the WISDM dataset, the optimal configuration uses 6-bit quantization with $d_\text{model}$ of 40, achieving 0.839 accuracy under integer-only inference, nearly identical to the FP32 result of 0.838. However, this gain in accuracy comes at the cost of increased computational overhead, reflected in a higher latency of 12.04 ms and energy consumption of 0.855 mJ. We attribute this to the reduced training set size and lower input dimensionality relative to~\cite{samanta2024optimizing}, which may drive the search toward larger models to compensate for limited data diversity. 

\subsection{Time-Series Anomaly Detection}

We conclude our evaluation with two threshold-based anomaly detection datasets.
ALFA\footnote{\url{https://github.com/castacks/alfa-dataset}} dataset captures multivariate UAV telemetry with 17 channels under fault injection (e.g., motor failure, sensor drift). Following~\cite{park2022model}, we predict 10 status variables one step ahead and apply smoothing-based residual thresholding, sweeping $\beta \in [0.749, 0.971]$ to select thresholds minimizing deviation from peak residuals.
SKAB\footnote{\url{https://github.com/waico/SKAB}} dataset contains 8-channel sensor readings from a water pump system under synthetic anomalies. We forecast the next flow rate and determine detection thresholds ($\text{Quantile}(r_\text{val}, 0.99)$) via a percentile-based rule on absolute residuals($r_\text{val}$).
All models are trained using MSE loss, and inference is performed by comparing predicted residuals against fixed thresholds computed on the validation set.

\begin{figure}[!htbp]
\vspace{-8pt}
    \centering
    \includegraphics[width=0.49\columnwidth]{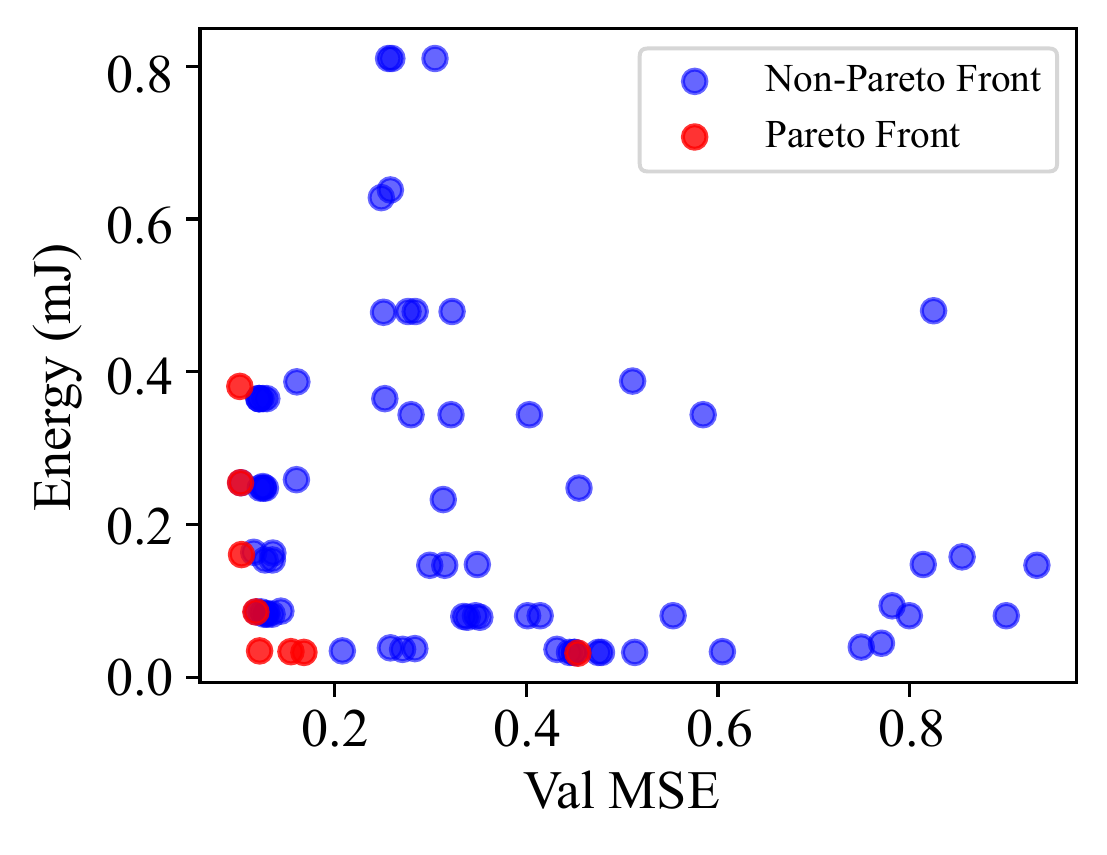}
    \includegraphics[width=0.49\columnwidth]{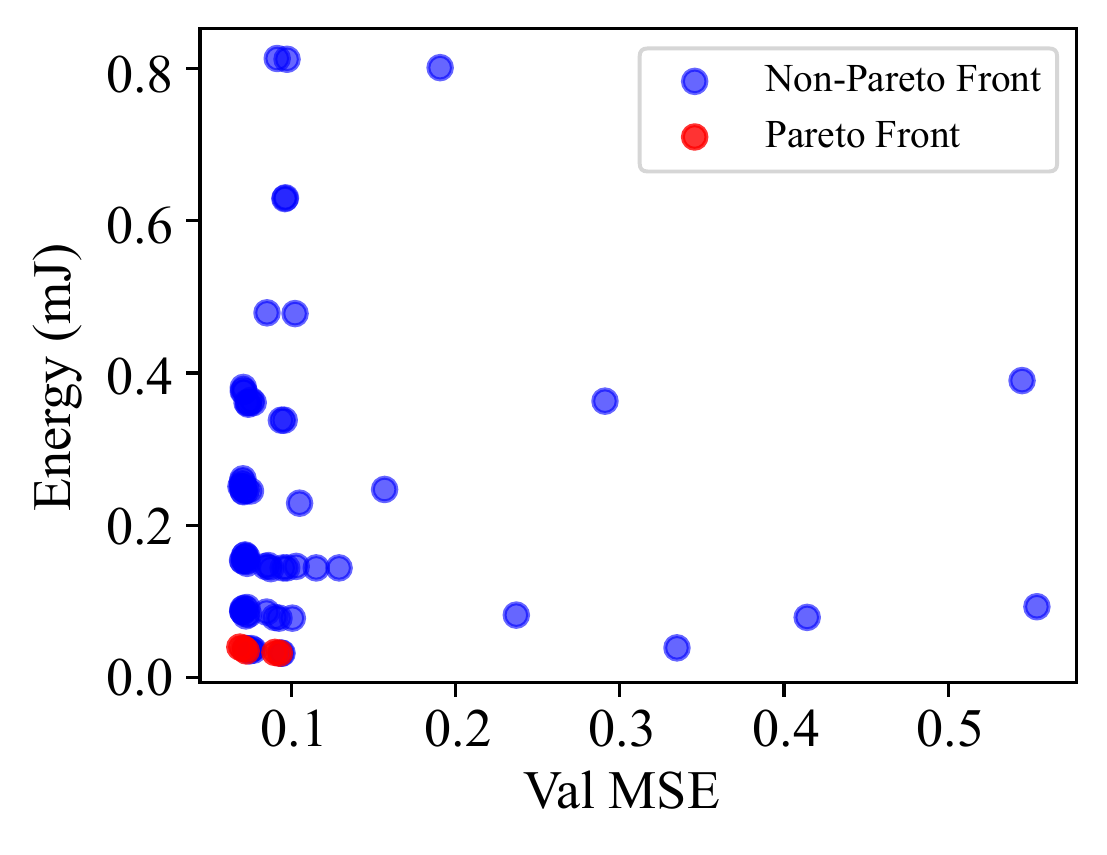}
    \caption{Pareto fronts on ALFA (left) and SKAB (right)}
    \label{fig:anomaly_pareto_amd}
\vspace{-14pt}
\end{figure}


Figure~\ref{fig:anomaly_pareto_amd} shows the Pareto fronts on XC7S15, with red points denoting optimal trade-offs between validation MSE and energy. Selected configurations are also reported in Table~\ref{tab:optuna_best_trial}.
On the ALFA dataset, the selected configuration ($d_\text{model}\!=\!8$, $bs\!=\!192$, $lr\!=\!1.2993 \times 10^{-3}$) achieves an F1-score of 0.889 under 4-bit quantization, representing a 3.68\% reduction compared to its FP32 counterpart (0.923). When compared to the stacked LSTM model from Park et al.~\cite{park2022model}, which achieves 0.96 F1-score, our model shows a 7.4\% accuracy gap. However, our model contains only 1106 parameters, with a total footprint of merely 0.91 KB, including both model parameters and intermediate buffer storage. This makes it 489$\times$ smaller than the 444.97 KB LSTM baseline. Our design achieves real-time performance, operating at 0.527\,ms latency with an energy cost of 0.033\,mJ per inference at 68\,mW. To our knowledge, this is the first work to demonstrate end-to-end quantized Transformer deployment on the ALFA dataset.

On the SKAB dataset, the selected model delivers an F1-score of 0.765 under 6-bit quantization, matching the FP32 baseline (0.766), while maintaining low deployment cost at only 0.154 mJ per inference. In comparison, Wen et al.~\cite{wen2022novel} achieve a higher F1-score of 0.929 on SKAB using a CNN-based adversarial autoencoder with dynamic thresholding. However, their approach prioritizes detection performance without explicit consideration of resource-constrained deployment. This underscores the strength of autoencoder-based methods for enhancing anomaly detection fidelity, while also highlighting the need for lightweight designs to enable efficient edge deployment.

\section{Conclusion and Future Work}
\label{sec:conclusion_future_work}

This paper introduces the first unified and fully automated deployment framework for Transformer-based time-series analysis on embedded FPGAs. The proposed system integrates QAT, modular VHDL code generation, and hardware-aware hyperparameter optimization to enable integer-only inference under strict resource constraints. We validate the framework across six public datasets on two embedded FPGA platforms, demonstrating successful deployment on AMD Spartan-7 and analyzing feasibility limits on Lattice iCE40. 

Beyond the directions discussed in Section~\ref{sec:results_eval}, future work will focus on improving the accuray of 4-bit quantized models through techniques such as knowledge distillation and refined quantization parameter calibration. In addition, we aim to extend the framework to support hybrid architectures that integrate Transformer with CNNs or LSTMs, enabling enhanced modeling of complex temporal dynamics.
\bibliographystyle{IEEEtran}
\bibliography{reference}
\end{document}